\documentclass{article} 
\usepackage{iclr2027_conference,times}

\usepackage{amsmath,amsfonts,bm}

\def\eqref#1{equation~\ref{#1}}

\def\1{\bm{1}}

\DeclareMathAlphabet{\mathsfit}{\encodingdefault}{\sfdefault}{m}{sl}
\SetMathAlphabet{\mathsfit}{bold}{\encodingdefault}{\sfdefault}{bx}{n}

\usepackage{hyperref}
\usepackage{url}
\usepackage{amsmath}
\usepackage{amssymb}
\usepackage{booktabs}
\usepackage{graphicx}
\usepackage{algorithm}
\usepackage{algpseudocode}
\usepackage{wrapfig}

\title{ATM: Why Latent World Models Can Fail to Plan}

\author{
Jiaheng Chen$^{1,2}$,
Tinghe Zhang$^{3,2}$,
Yucheng Xiao$^{1,2}$,
Xinyong Cai$^{4}$,
\\
Lan Yu$^{2}$,
Juncheng Bu$^{5}$,
Jiaxing Li$^{1,2}$,
Yunlong Wang$^{6,*}$
\\[2mm]
$^{1}$Harbin Institute of Technology, Shenzhen, China \\
$^{2}$Northeastern University,  China \\
$^{3}$Tsinghua University, Beijing, China \\
$^{4}$Tsinghua Shenzhen International Graduate School, Tsinghua University, Shenzhen, China \\
$^{5}$Peking University Shenzhen Graduate School, Shenzhen, China \\
$^{6}$Institute of Automation, Chinese Academy of Sciences, Beijing, China
\\[1mm]
$^{*}$Corresponding authors
}

\iclrfinalcopy 
\begin{document}

\maketitle

\fancyhead{}

\begin{abstract}
Latent world models can achieve accurate latent prediction yet still differ substantially in downstream planning performance. We argue that a key source of this discrepancy lies in the structure of action-induced latent transitions. We formalize action-identifiability through Bayes inverse risk, characterizing how much uncertainty about an action remains after observing the transition it induces. Model-predicted transitions can become highly self-decodable while encoding a domain-specific action relationship that fails to transfer to real environment transitions. We characterize this mismatch through cross-domain inverse transfer, instantiated as the Action-Consistency Transfer Matrix (ATM), a $2\times2$ inverse-risk matrix over real and predicted transition domains. Across TwoRoom, PushT, and OGBench-Cube, true-transition action-identifiability tracks downstream planning substantially better than standard prediction loss, while the full ATM reveals highly self-decodable yet cross-domain-inconsistent predicted transitions. Controlled interventions on the two domains further produce distinct transition structures and planning outcomes, supporting this decomposition. The same diagnostics also support lightweight model screening, reaching 98.81\% pairwise ranking accuracy for candidates separated by more than 5\% success.
\end{abstract}

\section{Introduction}
\label{sec:introduction}

Latent world models support goal-conditioned planning by rolling learned dynamics forward in representation space to evaluate candidate action sequences~\citep{ha2018worldmodels,hafner2019planet,hansen2022tdmpc}. Recent methods such as DINO-WM~\citep{zhou2024dinowm}, PLDM~\citep{sobal2025pldm}, and LeWM~\citep{maes2026lewm} further show that such planning does not require explicit future-observation reconstruction. Yet accurate latent prediction does not necessarily translate into reliable planning: models with favorable prediction quality can still differ substantially in downstream performance~\citep{grimm2020valueequivalence, wang_2026_decisionmetric}. Existing evaluation largely focuses on prediction accuracy or endpoint success, leaving unclear what properties of the learned representation actually make it useful for action selection. We argue that a missing piece lies in the structure of action-induced latent transitions. Although latent planning is typically formulated as goal matching, candidate actions affect the objective through transitions induced by the learned dynamics.

A planning-oriented representation should therefore preserve action effects so that different candidate actions remain distinguishable in latent transition space~\citep{pathak2017curiosity,brandfonbrener2023inverse,cui2024dynamo,han2026worldaction,zhang2026deltajepa}. We call this property \emph{action-identifiability}: the action inducing a latent transition should remain recoverable to a meaningful extent. Recent concurrent work has likewise emphasized transition-level action sensitivity, using action recovery or transition separation as representation-learning principles~\citep{boylan_2026_nogaussian,gan2026actswm}. If different actions induce weakly distinguishable or strongly entangled latent
changes, the representation may provide an unreliable interface for action
selection even when individual latent states are informative.
We show that action-identifiability admits a principled interpretation through
Bayes inverse risk, linking transition quality to the conditional uncertainty
of the action given the latent transition.

Action-identifiability characterizes action structure within a transition domain. A latent world model naturally induces two such domains: real encoded transitions from environment evolution and model-predicted transitions generated by the learned dynamics. Both domains may contain decodable action information yet encode actions through different relationships. In particular, predicted transitions can exhibit strong in-domain action consistency because the action is explicitly provided to the predictor; related planning-time methods test whether an inverse dynamics model can recover this conditioning action~\citep{seo2026acid}. This raises the question of whether this action–transition relationship is also preserved in real encoded transitions.

This motivates a cross-domain test of action structure. If real and predicted transitions preserve compatible action-conditioned relationships, an inverse rule learned in one domain should remain informative in the other; cross-domain degradation indicates that the action relationship changes between domains. We therefore introduce \textbf{cross-domain inverse transfer} and instantiate it as the \textbf{Action-Consistency Transfer Matrix (ATM)}, a $2\times2$ matrix formed by training inverse probes in each domain and evaluating them on both. Its diagonal entries characterize in-domain action-identifiability, while the off-diagonal entries measure cross-domain transfer of the action--transition relationship. Together, the four terms distinguish weak action structure from a qualitatively different failure mode: a domain-specific action relationship that is coherent within model rollouts but does not transfer to real transitions.

We validate this view on TwoRoom, PushT, and OGBench-Cube~\citep{park2025ogbench}. True-transition action-identifiability tracks planning substantially better than standard prediction loss, while the full ATM exposes models whose predicted transitions are highly self-decodable yet exhibit distinct cross-domain action structure. Controlled interventions further show that improving decodability in the two domains produces distinct transition structures and planning outcomes, supporting this distinction. Beyond diagnosis, the same quantities provide a lightweight signal for model screening. Together, these results show that latent planning failures reflect not only weak action information but also how it is represented and transferred between real and predicted transitions.

Our contributions are as follows:
\begin{enumerate}
    \item We identify \textbf{action-induced transition structure} as a key factor in latent planning and formalize action-identifiability through Bayes inverse risk.

    \item We show that predicted transitions can be highly self-decodable yet encode domain-specific action structure, and introduce \textbf{ATM} to characterize within-domain identifiability and cross-domain action transfer.

    \item We validate this decomposition through controlled interventions across three planning tasks and demonstrate its utility for model diagnosis and screening.
\end{enumerate}

\section{Related Work}

\paragraph{Latent world models for control and planning.}
Early work such as Ha and Schmidhuber's World Models~\citep{ha2018worldmodels} and PlaNet~\citep{hafner2019planet} showed that compact latent dynamics can support control from high-dimensional observations, while Dreamer~\citep{hafner2020dreamer,hafner2023dreamerv3} and TD-MPC-style methods~\citep{hansen2022tdmpc,hansen2023tdmpc2} further combined latent rollouts with policy optimization, value learning, or model predictive control. More recent reward-free or goal-conditioned methods, including DINO-WM~\citep{zhou2024dinowm}, PLDM~\citep{sobal2025pldm}, and LeWM~\citep{maes2026lewm}, plan directly in learned representation spaces with optimizers such as CEM~\citep{rubinstein1999crossentropy}. Recent analyses suggest that latent planning depends on properties beyond predictive accuracy. In particular, decision-metric alignment asks whether the latent metric used by MPC preserves decision-relevant rankings of candidate plans~\citep{wang_2026_decisionmetric}. ATM examines a complementary aspect of planning-oriented representations: the structure of action-induced latent transitions, including whether action effects are identifiable and whether their action--transition relationship transfers between real and predicted domains.

\paragraph{Evaluation and diagnostics for world models.}
Latent world models are commonly evaluated by downstream success, returns, or rollout-based planning performance~\citep{hafner2020dreamer,hansen2022tdmpc,zhou2024dinowm,maes2026lewm}. Such endpoint evaluation directly measures task performance but is planner-coupled, simulator-dependent, and often costly. Recent works also study post-hoc repairs or verifiers, including reachability-based planning costs~\citep{li2026trm}, action-reachability or action-consistency constraints~\citep{liu2026wav,wang2026groupactions}, and future--action compatibility for test-time selection~\citep{ruan2026futurecompatible}. A recent planning-time method also uses inverse-dynamics consistency within predicted rollouts as an action-selection signal~\citep{seo2026acid}. ATM directly examines whether an action--transition relationship learned in one domain remains valid in the other. This yields a \(2\times2\) cross-domain inverse-risk matrix without changing the planner or its objective.

\paragraph{Inverse dynamics and action-centric world modeling.}
Inverse dynamics has long been used to learn control-relevant representations, exploration signals, and action-conditioned features. Curiosity-driven exploration uses inverse dynamics to suppress uncontrollable visual variation~\citep{pathak2017curiosity}, while inverse-dynamics pretraining has been studied for imitation and visuomotor representation learning~\citep{brandfonbrener2023inverse}. Contrastive inverse dynamics has been incorporated into JEPA-style world-model training, using action discrimination over latent transitions as an anti-collapse signal~\citep{boylan_2026_nogaussian}. Recent action-sensitive world models encourage action recovery, action-conditioned transition separation, or action-aware dynamics~\citep{zhang2026deltajepa,gan2026actswm,han2026worldaction,tian2026starry}. These methods primarily investigate how to learn representations whose transitions remain sensitive to actions. ATM treats action-identifiability as a property of trained world models, formalizes it through Bayes inverse risk, and examines whether the action--transition relationship is preserved across real and predicted transition domains. AITS and AITS-P serve as controlled interventions to test this decomposition.

\begin{figure}[t]
    \centering
    \includegraphics[width=\linewidth]{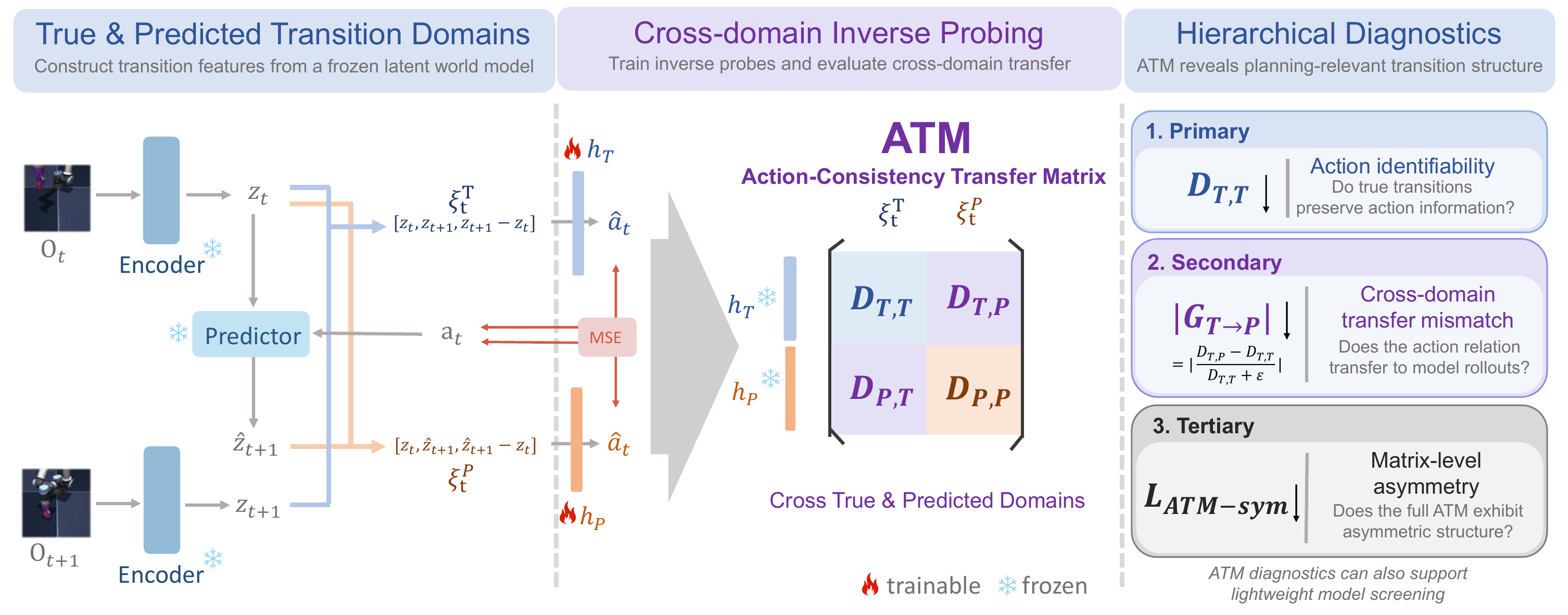}
    \vspace{-2em}
    \caption{
    Overview of ATM. A frozen latent world model induces true and predicted
    transition domains, where inverse probes are trained and evaluated cross-domain
    to form the $2\times2$ Action-Consistency Transfer Matrix. The resulting
    hierarchy uses $D_{T,T}$ for action-identifiability,
    $|G_{T\rightarrow P}|$ for transfer mismatch, and
    $L_{\mathrm{ATM\text{-}sym}}$ for matrix-level asymmetry.
    Snowflake and flame denote frozen and trainable modules, respectively.
    }
    \label{fig:atm_overview}
\end{figure}

\section{A TRANSITION-STRUCTURE VIEW OF LATENT PLANNING FAILURE}
\label{sec:method}

Figure~\ref{fig:atm_overview} summarizes our transition-structure view of latent planning failure. We first formalize action-identifiability through Bayes inverse risk. We then distinguish real encoded from model-predicted transitions and ask whether their conditional action relationships transfer across domains. This yields a cross-domain inverse-risk formulation and the Action-Consistency Transfer Matrix (ATM). Finally, we introduce controlled interventions to manipulate these properties independently.

\subsection{Latent Planning and Action-Identifiable Transitions}
\label{sec:action_identifiability}

We consider goal-conditioned planning with an action-conditioned latent world model.
An encoder $E_\theta$ maps observation $o_t$ to $z_t=E_\theta(o_t)$, and a dynamics
predictor produces $\hat z_{t+1}=F_\theta(z_t,a_t)$.
For multi-step planning, an optimizer such as CEM~\citep{rubinstein1999crossentropy} rolls the predictor forward to find an action sequence whose predicted terminal state best matches the latent goal $z_g=E_\theta(o_g)$:
\begin{equation}
    a^{*}_{t:t+H-1}
    =
    \arg\min_{a_{t:t+H-1}}
    d(\hat z_{t+H},z_g),
    \qquad
    \hat z_{t+k+1}=F_\theta(\hat z_{t+k},a_{t+k}).
\end{equation}

Although the planning objective is expressed through the predicted future state, candidate actions affect it through latent transitions induced by the learned dynamics. 
The key question is whether different actions induce distinguishable structure in latent transition space. 
We formalize this as \emph{action-identifiability}: how much information about an action remains recoverable from the latent transition it induces?

Consider an environment transition $(o_t,a_t,o_{t+1})$ and its latent feature $\xi_t=[z_t,z_{t+1},z_{t+1}-z_t]$, where $z_{t+1}=E_\theta(o_{t+1})$.
Let $\Xi$ and $A$ denote the corresponding random variables.
We characterize action-identifiability by the minimum squared inverse-prediction risk. Under squared loss, the Bayes-optimal decoder is $h^*(\xi)=\mathbb{E}[A\mid\Xi=\xi]$, yielding
\begin{equation}
    R^*
    =
    \inf_h
    \mathbb{E}\!\left[\|h(\Xi)-A\|_2^2\right]
    =
    \mathbb{E}\!\left[
        \operatorname{tr}\operatorname{Var}(A\mid\Xi)
    \right].
\end{equation}

Lower inverse risk means that the latent transition leaves less uncertainty about the inducing action, whereas higher risk indicates weaker action-identifiable structure.
$R^*$ is a property of the joint transition–action distribution and is independent of any particular inverse model. This gives action-identifiability a direct statistical interpretation.
In practice, we estimate this quantity with the lightweight probes in Section~\ref{sec:cross_domain_atm}.

Irreducible inverse ambiguity can arise when multiple actions induce equivalent transition consequences. In such cases, cross-domain preservation of the action–transition relationship is the relevant criterion.

Action-identifiability, however, is a within-domain property: it characterizes recoverable action information within a transition domain, but not whether the same action--transition relationship is shared across domains.

\subsection{Real and Predicted Transition Domains}
\label{sec:transition_domains}

A latent world model induces two transition domains from the same tuple $(o_t,a_t,o_{t+1})$.
The first is the \emph{real encoded transition domain}, obtained by encoding the observed next state.
The second is the \emph{predicted transition domain}, obtained by applying the learned dynamics
under the same current latent state and action.
We represent them as
\begin{equation}
    \xi_t^T
    =
    [\,z_t,\;z_{t+1},\;z_{t+1}-z_t\,],
    \qquad
    \xi_t^P
    =
    [\,z_t,\;\hat z_{t+1},\;\hat z_{t+1}-z_t\,],
\end{equation}
where $z_{t+1}=E_\theta(o_{t+1})$ and
$\hat z_{t+1}=F_\theta(z_t,a_t)$.
The two features share the same current latent state and action, differing only in whether the next latent state comes from environment evolution or the learned predictor.

The two domains need not encode actions through the same relationship.
Let $\mathcal{D}_T$ and $\mathcal{D}_P$ denote their joint transition--action distributions.
They induce different conditional action relationships, $p_T(a\mid\xi^T)$ and $p_P(a\mid\xi^P)$.
Under squared loss, their Bayes-optimal inverse rules are
\begin{equation}
    h_T^*(\xi)
    =
    \mathbb{E}_{\mathcal{D}_T}[A\mid\Xi^T=\xi],
    \qquad
    h_P^*(\xi)
    =
    \mathbb{E}_{\mathcal{D}_P}[A\mid\Xi^P=\xi].
\end{equation}
By Eq.~(2), the in-domain Bayes risks characterize action-identifiability within each domain.
Low inverse risk in both domains, however, does not imply a shared action--transition relationship.

This distinction is particularly important for model-predicted transitions.
Since the action is an explicit predictor input, predicted transitions may encode action information in a model-specific form that is highly decodable within the model's rollout space.
Low predicted-domain inverse risk therefore shows only that actions are recoverable from model-generated transitions; it does not imply the same action--transition relationship in real encoded transitions.

This separates \emph{in-domain action decodability} from
\emph{cross-domain action transfer}.
If the two domains preserve compatible action-conditioned structure, an inverse relationship learned in one should remain informative in the other.
Conversely, cross-domain degradation indicates that the action--transition relationship changes between real and predicted dynamics.
We operationalize this transfer principle with the Action-Consistency Transfer Matrix in the next
section.

\vspace{4pt}

\subsection{Cross-Domain Inverse Transfer and ATM}
\label{sec:cross_domain_atm}

We operationalize the domain-specific Bayes inverse rules from Section~\ref{sec:transition_domains} with a shared lightweight probe family.
For each world model, the encoder and predictor remain frozen; we train probes $h_T$ and $h_P$ on $\mathcal{D}_T$ and $\mathcal{D}_P$, respectively.
Both probes use the same two-layer MLP with LayerNorm and GELU; token- or patch-level latents are mean-pooled before feature construction.

For $i,j\in\{T,P\}$, we define
\begin{equation}
\begin{aligned}
    h_i
    &\approx
    \arg\min_h
    \mathbb{E}_{(\xi,a)\sim\mathcal{D}_i}
    \left[\|h(\xi)-a\|_2^2\right], \\
    D_{i,j}
    &=
    \mathbb{E}_{(\xi,a)\sim\mathcal{D}_j}
    \left[\|h_i(\xi)-a\|_2^2\right],
    \qquad
    \mathbf{A}_{\mathrm{ATM}}
    =
    \begin{bmatrix}
        D_{T,T} & D_{T,P} \\
        D_{P,T} & D_{P,P}
    \end{bmatrix}.
\end{aligned}
\end{equation}

Rows of $\mathbf{A}_{\mathrm{ATM}}$ denote the training domain and columns the evaluation domain.
Diagonal entries measure in-domain action decodability, while off-diagonal entries measure transfer of the learned action--transition relationship.

Specifically, $D_{T,T}$ approximates true-domain action-identifiability, while $D_{T,P}$ tests whether this inverse relationship transfers to predicted transitions.
Conversely, $D_{P,P}$ measures predicted-domain self-decodability, while $D_{P,T}$ tests whether this relationship transfers to real transitions.
Thus, $D_{P,P}\ll D_{P,T}$ exposes a \emph{predicted-domain-specific action code}: actions are easy to recover in the model's rollout space, but the same decoding relationship fails on real transitions.

We summarize directional cross-domain transfer with the relative gaps
\begin{equation}
\begin{aligned}
    G_{T\rightarrow P}
    &=
    \frac{D_{T,P}-D_{T,T}}
         {D_{T,T}+\epsilon},
    &
    G_{P\rightarrow T}
    &=
    \frac{D_{P,T}-D_{P,P}}
         {D_{P,P}+\epsilon}.
\end{aligned}
\end{equation}
Here $\epsilon$ is a small stability constant.
$G_{T\rightarrow P}$ measures the relative change when the true-domain inverse relationship transfers to predicted transitions, and $G_{P\rightarrow T}$ the reverse.
Because cross-domain evaluation can occasionally improve inverse prediction, we use magnitudes to measure mismatch and signed values to analyze transfer direction.

ATM is thus an empirical cross-domain inverse-risk matrix whose diagonal entries
quantify within-domain action-identifiability and off-diagonal entries
cross-domain transfer.
The full matrix can distinguish weak action information from domain-specific
action coding---a distinction that no single inverse loss can reveal.

\subsection{Matrix-Level Asymmetry}
\label{sec:matrix_imbalance}

The directional gaps characterize directional transfer mismatch, while the full ATM may also exhibit asymmetry between the real and predicted domains.
We summarize this behavior with a compact matrix-level statistic, retaining individual entries and directional gaps as the primary diagnostics.

We first measure the discrepancy between the two in-domain inverse risks as
\begin{equation}
\begin{aligned}
    I_{\mathrm{diag}}
    &=
    \frac{|D_{T,T}-D_{P,P}|}
    {\frac{1}{2}(D_{T,T}+D_{P,P})+\epsilon}, \\
    L_{\mathrm{raw}}
    &=
    I_{\mathrm{diag}}
    +
    \left|G_{T\rightarrow P}-G_{P\rightarrow T}\right|, \\
    L_{\mathrm{ATM\text{-}sym}}
    &=
    \frac{\log(1+L_{\mathrm{raw}})}
    {1+\log(1+L_{\mathrm{raw}})}.
\end{aligned}
\end{equation}

Here, $I_{\mathrm{diag}}$ measures in-domain decodability discrepancy, while the second term measures directional transfer asymmetry.
Logarithmic compression limits extreme relative gaps, and the final monotonic mapping bounds $L_{\mathrm{ATM\text{-}sym}}$ to $[0,1)$.
Near-zero values indicate similar in-domain decodability and relatively symmetric transfer; larger values indicate stronger matrix-level asymmetry.

$L_{\mathrm{ATM\text{-}sym}}$ is a tertiary rather than standalone diagnostic.
Lower asymmetry is preferable when the primary ATM quantities are comparable, but should not override large differences in action-identifiability or directional transfer.
We therefore use it as a tie-breaking summary, with $D_{T,T}$ and $|G_{T\rightarrow P}|$ as the primary diagnostics.

\subsection{Controlled Interventions on Transition Structure}
\label{sec:controlled_interventions}

The diagnostics above characterize transition structure after training.
To test whether the identified properties respond differently to targeted domain-specific changes, we introduce two training-time interventions.
These interventions selectively increase action decodability in real or predicted transitions while leaving the planner unchanged.

Let $h_{\psi}^{T}$ and $h_{\psi}^{P}$ denote training-time inverse heads for real and predicted transition features. We define
\begin{equation}
\begin{aligned}
    \mathcal{L}_{T}
    &=
    \mathbb{E}\!\left[
        \|h_{\psi}^{T}(\xi_t^T)-a_t\|_2^2
    \right], \\
    \mathcal{L}_{P}
    &=
    \mathbb{E}\!\left[
        \|h_{\psi}^{P}(\xi_t^P)-a_t\|_2^2
    \right], \\
    \mathcal{L}
    &=
    \mathcal{L}_{\mathrm{WM}}
    +
    \alpha_T \mathcal{L}_{T}
    +
    \alpha_P \mathcal{L}_{P}.
\end{aligned}
\end{equation}

We call the true-transition intervention ($\alpha_T>0,\alpha_P=0$) \emph{Action-Identifiable Transition Supervision} (AITS), and the predicted-transition intervention ($\alpha_T=0,\alpha_P>0$) AITS-P.
We also consider their combination by enabling both terms.
AITS encourages action-identifiable real transitions, whereas AITS-P encourages action decodability in the predicted transition space.

Neither intervention directly optimizes cross-domain transfer.
Their comparison therefore tests whether increasing in-domain action decodability in different domains yields the same cross-domain structure.
In particular, stronger predicted-domain decodability does not by construction require the learned inverse relationship to remain valid on real transitions.

The training-time inverse heads are discarded after training.
For ATM evaluation, we freeze the encoder and predictor and train fresh probes $h_T$ and $h_P$ from scratch.
This ensures that ATM reflects the learned world model rather than the auxiliary intervention heads.

\section{Experiments}
\label{sec:experiments}

We organize our experiments around the diagnostic hierarchy above.
First, we test whether true-transition action-identifiability captures planning-relevant variation beyond prediction loss and whether cross-domain quantities provide further refinement.
Second, we inspect the full ATM for failure modes invisible to in-domain decodability alone.
Third, controlled interventions test whether targeted changes in the real and predicted domains yield distinct cross-domain structures and planning outcomes.
Finally, we evaluate the same diagnostics for lightweight model screening.

\subsection{Experimental Setup}
\label{sec:exp_setup}

\paragraph{Tasks.}
We evaluate three goal-conditioned planning tasks: TwoRoom, PushT, and OGBench-Cube~\citep{park2025ogbench}.
TwoRoom is 2D navigation, PushT continuous-control 2D manipulation, and OGBench-Cube more complex 3D manipulation.
Together, they span increasingly challenging transition structures, from navigation to contact-rich manipulation and visually complex 3D control.

\paragraph{Models.}
Our controlled experiments use LeWM-style~\citep{maes2026lewm} latent world models.
From a shared LeWM baseline, we compare AITS, AITS-P, and their combination, selectively intervening on action decodability in the real domain, predicted domain, or both.
We also include DINO-WM~\citep{zhou2024dinowm} checkpoints as an external family to test whether similar real--predicted transition structure arises beyond these interventions.
All variants within each task use the same CEM planning protocol.

\subsection{A Hierarchy of Planning-Relevant Transition Diagnostics}
\label{sec:action_identifiability_results}

We first test whether action-identifiability captures planning-relevant variation beyond prediction loss and whether cross-domain transfer adds discrimination.
We focus on OGBench-Cube, whose less-saturated success rates make it more suitable for studying continuous relationships between transition diagnostics and planning performance.

Table~\ref{tab:identifiability_corr} compares latent prediction loss, true-transition action-identifiability, and a simple combination additionally penalizing true-to-predicted transfer mismatch.
All quantities are oriented so higher values indicate better quality.

\begin{table}[t]
\centering
\caption{Correlation with downstream planning success on OGBench-Cube.
Higher is better. Here, $\lambda$ is fitted within the same task on the candidate pool.}
\label{tab:identifiability_corr}
\begin{tabular}{lcc}
\toprule
Score & Spearman $\rho$ & Pearson $r$ \\
\midrule
$-\mathcal{L}_{\mathrm{pred}}$ 
& 0.4983 & 0.3927 \\

$-D_{T,T}$ 
& 0.8130 & \textbf{0.7015} \\

$-D_{T,T}-\lambda |G_{T\rightarrow P}|$
& \textbf{0.8239} & 0.7006 \\
\bottomrule
\end{tabular}
\end{table}

The largest gain comes from replacing prediction loss with true-transition
action-identifiability: Spearman correlation increases from 0.4983 to 0.8130 and
Pearson correlation from 0.3927 to 0.7015.
This supports $D_{T,T}$ as the primary planning-relevant diagnostic.
Adding the true-to-predicted transfer term further improves rank correlation to 0.8239,
while leaving linear correlation nearly unchanged, suggesting that cross-domain structure
provides secondary discrimination beyond in-domain identifiability.

This supports the hierarchy in Section~3: $D_{T,T}$ captures the dominant planning-relevant variation, while $|G_{T\rightarrow P}|$ adds discrimination when in-domain action-identifiability is insufficient.
Higher-order matrix structure is analyzed next.

\begin{figure}[t]
    \centering
    \vspace{-4pt}
    \includegraphics[width=\linewidth]{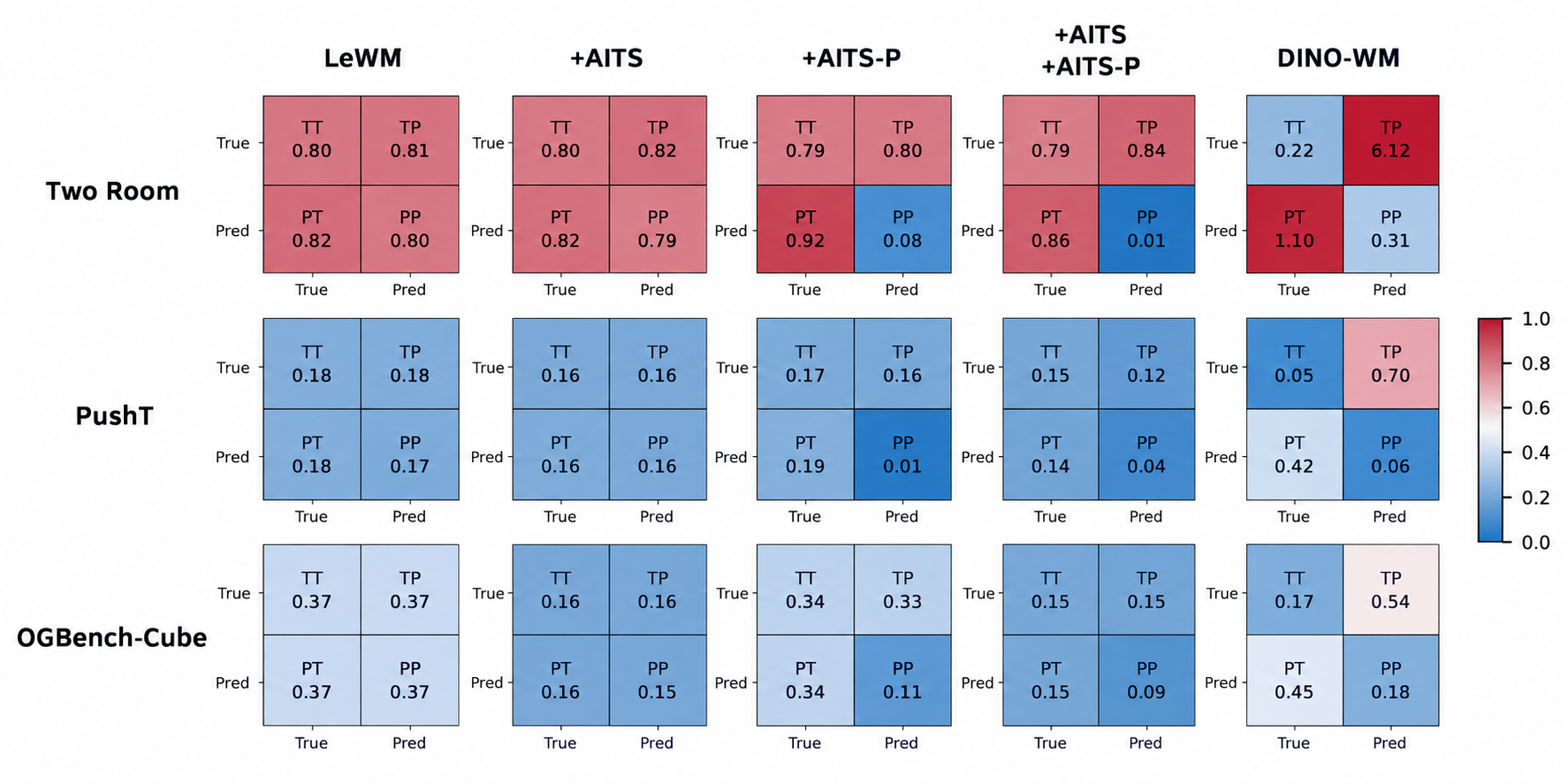}
    \vspace{-8pt}
    \caption{
    Full ATM matrices across tasks and model variants.
    Rows denote training domains and columns evaluation domains.
    Low diagonal loss indicates strong in-domain action decodability, while diagonal-to-cross-domain gaps reveal transfer mismatch.
    Predicted-transition supervision can yield low $D_{P,P}$ but much larger $D_{P,T}$, exposing a self-decodable yet domain-specific action code.
    }
    \label{fig:atm_matrices}
\end{figure}

\subsection{Cross-Domain Transfer Reveals Hidden Failure Modes}
\label{sec:cross_domain_results}

Having established true-transition action-identifiability as the primary diagnostic, we next examine what cross-domain structure reveals beyond in-domain decodability.
Figure~\ref{fig:atm_matrices} shows the full ATM matrices across model variants, families, and all three tasks.

A recurring pattern is that strong in-domain decodability does not imply cross-domain transfer of the same action--transition relationship.
Several predicted-transition-supervised models achieve low $D_{P,P}$ but substantially larger $D_{P,T}$: actions are highly decodable within predicted rollouts, yet the inverse relationship fails on real transitions.
This is the predicted-domain-specific action structure characterized in Section~\ref{sec:cross_domain_atm}.

The full ATM therefore reveals distinctions invisible to either diagonal alone.
Models with similar in-domain inverse risks can differ markedly in cross-domain transfer, while highly decodable predicted transitions can still rely on domain-specific action codes.
Hence, cross-domain quantities provide secondary discrimination beyond true-transition action-identifiability.

This pattern appears across tasks and model configurations.
DINO-WM~\citep{zhou2024dinowm} checkpoints further show that real--predicted transition mismatch arises in an independently trained world-model family beyond our controlled transition-supervision objectives.

Together, these observations support ATM's central distinction:
\emph{action information can be highly decodable within a transition domain without the underlying action--transition relationship transferring across real and predicted dynamics.}
We next test this distinction directly through controlled interventions.

\begin{table*}[t]
\centering
\caption{
Controlled intervention analysis across three tasks.
$D_{T,T}$ and $D_{P,P}$ measure in-domain action decodability,
$|G_{T\to P}|$ measures true-to-predicted transfer mismatch, and
$L_{\mathrm{ATM\text{-}sym}}$ summarizes higher-order matrix asymmetry.
All ATM diagnostic values are scaled by $100$ for readability.
Lower $L_{\mathrm{ATM\text{-}sym}}$ is preferable as a tertiary criterion when
the primary diagnostics are comparable.
Success denotes downstream planning success (\%).
}
\label{tab:intervention_analysis}
\begin{tabular}{llccccc}
\toprule
Task & Variant
& $D_{T,T}\downarrow$
& $D_{P,P}\downarrow$
& $|G_{T\to P}|\downarrow$
& $L_{\mathrm{ATM\text{-}sym}}\downarrow$
& Success$\uparrow$ \\
\midrule

TwoRoom
& LeWM
& 80.26 & 80.00 & 1.28 & 0.95 & 87 \\

TwoRoom
& + AITS
& 79.56 & 79.00 & 2.50 & 1.15 & 92 \\

TwoRoom
& + AITS-P
& 79.36 & 8.00 & 0.68 & 72.43 & 94 \\

TwoRoom
& + AITS + AITS-P
& 79.13 & 1.00 & 6.09 & 81.15 & 92 \\

\midrule

PushT
& LeWM
& 18.29 & 17.00 & 3.72 & 16.94 & 96 \\

PushT
& + AITS
& 16.23 & 16.00 & 1.22 & 7.80 & 100 \\

PushT
& + AITS-P
& 17.34 & 1.00 & 10.56 & 72.83 & 100 \\

PushT
& + AITS + AITS-P
& 15.34 & 4.00 & 24.99 & 62.88 & 98 \\

\midrule

OGBench-Cube
& LeWM
& 36.65 & 37.00 & 0.45 & 1.07 & 74 \\

OGBench-Cube
& + AITS
& 15.87 & 15.00 & 0.96 & 7.44 & 84 \\

OGBench-Cube
& + AITS-P
& 33.77 & 11.00 & 1.60 & 58.13 & 78 \\

OGBench-Cube
& + AITS + AITS-P
& 15.40 & 9.00 & 4.57 & 45.03 & 84 \\

\bottomrule
\end{tabular}
\vspace{-14pt}
\end{table*}

\subsection{Controlled Interventions Separate Decodability from Transfer}
\label{sec:intervention_results}

We next use the controlled interventions from Section~\ref{sec:controlled_interventions} to test whether increasing action decodability in different domains yields the same learned structure.
Neither AITS nor AITS-P directly optimizes cross-domain transfer: AITS targets real-transition decodability, whereas AITS-P targets predicted-transition decodability.

Table~\ref{tab:intervention_analysis} shows distinct ATM signatures across the interventions.
AITS improves true-transition action-identifiability and planning across all three tasks, increasing success from $87\%$ to $92\%$ on TwoRoom, $96\%$ to $100\%$ on PushT, and $74\%$ to $84\%$ on OGBench-Cube.
On OGBench-Cube, $D_{T,T}$ decreases from $36.65$ to $15.87$.

AITS-P produces a different structure.
It can sharply reduce $D_{P,P}$, making predicted transitions highly self-decodable without a correspondingly transferable action relationship.
Figure~\ref{fig:atm_matrices} shows a large discrepancy between $D_{P,P}$ and $D_{P,T}$, revealing a predicted-domain-specific action code.
Thus, predicted-domain decodability alone does not ensure compatibility with real transitions.

Planning results further clarify the hierarchy.
AITS-P remains effective on the simpler TwoRoom and PushT tasks despite large matrix asymmetry, but on OGBench-Cube improves success only from $74\%$ to $78\%$, versus $84\%$ under AITS.
Applying both interventions likewise does not consistently outperform AITS alone.
These results support ATM's distinction: in-domain action decodability is useful, but where it arises---and whether the corresponding relationship transfers---can matter for planning, especially with more demanding transition structure.

\subsection{Practical Model Screening with ATM}
\label{sec:model_screening}

Finally, we ask whether the transition-level quantities above can also support
practical model selection.
Following Section~\ref{sec:action_identifiability_results}, we use a simple
task-specific score combining true-transition action-identifiability with
true-to-predicted transfer mismatch:
\begin{equation}
    S_{\mathrm{ATM}}
    =
    -D_{T,T}
    -
    \lambda |G_{T\rightarrow P}|,
\end{equation}
where $\lambda$ is fitted within the same task.
We use this score only for relative ranking rather than as a calibrated predictor
of absolute planning success.

Table~\ref{tab:screening_ranking} reports pairwise ranking on OGBench-Cube, whose candidates span a broad range of planning performance.
A pair is correct when $S_{\mathrm{ATM}}$ agrees with the simulator-based success ordering.

\begin{table}[t]
\centering
\caption{
Pairwise model-ranking accuracy on OGBench-Cube using
$S_{\mathrm{ATM}}=-D_{T,T}-\lambda|G_{T\to P}|$.
Margins exclude pairs with only small differences in simulator-based success.
}
\label{tab:screening_ranking}
\begin{tabular}{lccc}
\toprule
Success margin
& $n_{\mathrm{pairs}}$
& Ranking acc.
& Weighted acc. \\
\midrule
All pairs & 1865 & 82.84 & 92.61 \\
$>2\%$    & 1338 & 91.48 & 95.14 \\
$>5\%$    & 841  & 98.81 & 99.09 \\
$>10\%$   & 123  & 100.00 & 100.00 \\
\bottomrule
\end{tabular}

\vspace{-4pt}
\end{table}

The score correctly ranks $82.84\%$ of all pairs and becomes more reliable as the performance gap increases.
For gaps above $5\%$, accuracy reaches $98.81\%$; all pairs above $10\%$ are ranked correctly.
Thus, the same diagnostics provide a lightweight signal for selecting meaningfully different candidate models.

We further compare LeWM-style and DINO-WM candidates across families.
Transferring the coefficient fitted on LeWM-style models yields $68.54\%$ accuracy on LeWM--DINO-WM pairs.
Refitting the single coefficient on the mixed pool raises cross-family accuracy to $89.90\%$, with $83.10\%$ over all pooled pairs.
This suggests that the underlying ATM quantities remain informative across model
families, while their relative scales may require lightweight calibration.

Model screening is a practical consequence rather than the defining purpose of ATM: once latent transitions are cached, ATM takes only 3--5 seconds per checkpoint in our setup, versus minutes to hours for simulator-based CEM evaluation. The core contribution remains separating in-domain action-identifiability from cross-domain transfer.

\section{Conclusion}

We study why latent world models with similar prediction quality can exhibit different planning performance. We identify action-induced transition structure as a key factor: true-transition action-identifiability provides the primary signal, while cross-domain transfer reveals failure modes. We formalize these properties through Bayes inverse risk and jointly diagnose them with the Action-Consistency Transfer Matrix (ATM). Across three tasks, action-identifiability tracks planning substantially better than prediction loss, while the full ATM exposes highly self-decodable yet domain-specific predicted transitions; controlled interventions produce distinct transition structures and planning outcomes. 
These results identify two sources of latent planning failure: weak action information and failure of predicted-transition action structure to transfer to real dynamics.

\section{Discussion}

ATM also provides a lens on desirable properties of planning-oriented latent representations.
Our results suggest a hierarchy: strong true-transition action-identifiability is primary, while cross-domain transfer and matrix balance provide refinement when identifiability is comparable.
World-model families exhibit different trade-offs along this hierarchy, suggesting that effective latent representations should first preserve clear action-induced structure in real transitions and then maintain compatible action relationships in predicted transitions.

\subsection*{AI use statement}

We used generative AI tools to provide feedback on research methodology and
experimental presentation, assist in interpreting experimental results, support
code implementation and debugging, and assist with manuscript drafting and editing.
We additionally used generative AI for literature search and summarization,
identifying relevant work, and refining figures and paper structure.
All AI-assisted suggestions were independently reviewed by the authors; cited
literature was verified against original sources, AI-assisted code was reviewed
and tested against the intended experimental behavior, reported numerical results
were checked against our experimental outputs, and all methodological decisions
and final claims were made by the authors.
We take responsibility for the final content of this work, including text, claims,
code, and artifacts produced with the aid of generative AI.

\subsection*{Reproducibility statement}

We provide detailed experimental and implementation information in the appendix.
Appendix~A specifies the evaluated tasks and world models, ATM transition
construction and probe protocol, and the controlled-intervention setup.
The theoretical derivations underlying the Bayes inverse-risk formulation are
provided in Appendix~B.
Additional screening and cross-family evaluation details are provided in
Appendix~C, and computational-cost measurements in Appendix~D.
Code will be released upon acceptance.




\bibliography{references}

\bibliographystyle{iclr2027_conference}

\appendix

\section{Experimental and Implementation Details}
\label{app:experimental_details}

\subsection{Tasks and Evaluated World Models}
\label{app:tasks_models}

We evaluate ATM on the same three goal-conditioned planning tasks used in
the main experiments: TwoRoom, PushT, and OGBench-Cube~\citep{park2025ogbench}.
TwoRoom is a 2D navigation task, PushT is a continuous-control 2D manipulation
task, and OGBench-Cube is a visually more complex 3D manipulation task.
Together, they span increasingly challenging transition structures, from
navigation to contact-rich manipulation and 3D control.

Our controlled experiments use LeWM-style~\citep{maes2026lewm} latent world
models. Starting from the same baseline, we consider AITS, AITS-P, and their
combination, which intervene on action decodability in the true transition
domain, the predicted transition domain, or both. We additionally include
DINO-WM~\citep{zhou2024dinowm} checkpoints as an independently trained
world-model family. All variants within each task are evaluated under the
same CEM planning protocol used for the corresponding comparisons in the
main paper.

\subsection{ATM Transition Features and Probe Protocol}
\label{app:probe_protocol}

For each offline transition tuple $(o_t,a_t,o_{t+1})$, the frozen world model
produces the true and predicted transition features
\begin{align}
\xi_t^T &=
[z_t,z_{t+1},z_{t+1}-z_t], \\
\xi_t^P &=
[z_t,\hat z_{t+1},\hat z_{t+1}-z_t],
\end{align}
where
\[
z_t = E_\theta(o_t), \qquad
z_{t+1}=E_\theta(o_{t+1}), \qquad
\hat z_{t+1}=F_\theta(z_t,a_t).
\]

The two transition domains are therefore constructed from the same underlying
transition tuple and share the same current latent state and action; they differ
only in whether the next latent state is obtained from actual environment
evolution or from the learned predictor.

For vector-valued latent states, the latent representation is used directly.
For token- or patch-level representations, we first apply mean pooling over
tokens before constructing the transition feature. This gives a fixed-size
representation across the evaluated world-model families.

Both ATM probes use the same lightweight two-layer MLP architecture with
LayerNorm and GELU activation,
\[
h(\xi)
=
W_2\,\sigma\!\left(
    \mathrm{LN}(W_1\xi+b_1)
\right)+b_2.
\]
For every evaluated world-model checkpoint, the encoder and dynamics predictor
remain frozen, and the true-domain probe $h_T$ and predicted-domain probe $h_P$
are trained from scratch. Each probe is then evaluated on both transition
domains to form
\[
\mathbf{A}_{\mathrm{ATM}}
=
\begin{bmatrix}
D_{T,T} & D_{T,P} \\
D_{P,T} & D_{P,P}
\end{bmatrix}.
\]
This separation ensures that ATM measures action information accessible in the
frozen transition representation rather than the performance of a head trained
jointly with the world model.

\subsection{Controlled-Intervention Protocol}
\label{app:intervention_protocol}

The controlled interventions used in the main paper modify the domain in which
action decodability is explicitly encouraged. With training-time inverse heads
$h_\psi^T$ and $h_\psi^P$, we use
\begin{align}
\mathcal{L}_T
&=
\mathbb{E}
\left[
\|h_\psi^T(\xi_t^T)-a_t\|_2^2
\right],\\
\mathcal{L}_P
&=
\mathbb{E}
\left[
\|h_\psi^P(\xi_t^P)-a_t\|_2^2
\right],
\end{align}
and
\[
\mathcal{L}
=
\mathcal{L}_{\mathrm{WM}}
+
\alpha_T\mathcal{L}_T
+
\alpha_P\mathcal{L}_P.
\]

The variants correspond to the following intervention structure:
\begin{table}[ht]
\centering
\caption{
Controlled interventions used in the experiments.
The table indicates which transition domain receives explicit
action-decoding supervision.
}
\label{tab:app_interventions}
\begin{tabular}{lcc}
\toprule
Variant & $\alpha_T$ & $\alpha_P$ \\
\midrule
LeWM baseline & $0$ & $0$ \\
AITS & $>0$ & $0$ \\
AITS-P & $0$ & $>0$ \\
AITS + AITS-P & $>0$ & $>0$ \\
\bottomrule
\end{tabular}
\end{table}

The training-time inverse heads are discarded after world-model training.
ATM evaluation always uses newly initialized post-hoc probes on the frozen
world model. Consequently, neither AITS nor AITS-P directly optimizes the
cross-domain ATM quantities evaluated after training.

\section{Additional Theoretical Details}
\label{app:theory}

\subsection{From Latent Planning to Action-Induced Transitions}
\label{app:planning_chain}

For goal-conditioned latent planning, a planner such as CEM searches for an
action sequence by rolling the learned dynamics forward:
\[
\hat z_{t+k+1}
=
F_\theta(\hat z_{t+k},a_{t+k}),
\qquad
k=0,\ldots,H-1,
\]
and selecting
\[
a^*_{t:t+H-1}
=
\arg\min_{a_{t:t+H-1}}
d(\hat z_{t+H},z_g).
\]
Equivalently, the planning objective can be viewed as a function of the action
sequence,
\[
J_\theta(a_{t:t+H-1})
=
d\!\left(
F_\theta^{(H)}(z_t,a_{t:t+H-1}),
z_g
\right).
\]

Thus, although the endpoint objective is expressed as latent goal matching,
candidate actions affect that objective through the latent transitions induced
by the learned dynamics. Locally, for a perturbation $\delta a$,
\[
F_\theta(z_t,a_t+\delta a)
\approx
F_\theta(z_t,a_t)
+
J_aF_\theta(z_t,a_t)\delta a,
\]
where $J_aF_\theta$ denotes the action Jacobian of the latent dynamics.
This motivates examining whether action-induced latent transitions retain
distinguishable action information.

ATM does not attempt to estimate this Jacobian directly, nor is inverse
decodability an algebraic consequence of the CEM objective. Instead, it provides
a model-agnostic diagnostic of the transition interface through which candidate
actions affect latent planning.

\subsection{Bayes Inverse Risk}
\label{app:bayes_risk}

Let $\Xi$ denote a latent transition feature and $A$ the corresponding action.
For any inverse predictor $h$, the squared inverse risk is
\[
R(h)
=
\mathbb{E}
\left[
\|A-h(\Xi)\|_2^2
\right].
\]
Conditioning on $\Xi$ gives the standard decomposition
\begin{align}
\mathbb{E}
\left[
\|A-h(\Xi)\|_2^2
\mid \Xi
\right]
&=
\mathbb{E}
\left[
\|A-\mathbb{E}[A\mid\Xi]\|_2^2
\mid \Xi
\right]
\nonumber\\
&\quad+
\|\mathbb{E}[A\mid\Xi]-h(\Xi)\|_2^2 .
\end{align}
The second term is non-negative and vanishes for
\[
h^*(\xi)
=
\mathbb{E}[A\mid\Xi=\xi].
\]
Therefore,
\[
R^*
=
\inf_h R(h)
=
\mathbb{E}
\left[
\operatorname{tr}\operatorname{Var}(A\mid\Xi)
\right].
\]

This identity gives action-identifiability a population-level interpretation.
A lower Bayes inverse risk means that less uncertainty about the action remains
after observing the induced transition. The lightweight ATM probes provide an
empirical approximation to this quantity for a frozen world model.

Importantly, irreducible inverse ambiguity does not necessarily indicate a
deficient representation. If multiple actions induce equivalent transition
consequences, the corresponding action labels need not be uniquely recoverable.
In such cases, the relevant question is whether the same action--transition
relationship, including its intrinsic ambiguity, is preserved across the true
and predicted transition domains.

\subsection{Why Cross-Domain Transfer Is Needed}
\label{app:cross_domain_theory}

The true and predicted transition domains induce potentially different
conditional action relationships,
\[
p_T(a\mid\xi^T)
\qquad\text{and}\qquad
p_P(a\mid\xi^P),
\]
with Bayes-optimal inverse rules
\[
h_T^*(\xi)
=
\mathbb{E}_{D_T}[A\mid\Xi^T=\xi],
\qquad
h_P^*(\xi)
=
\mathbb{E}_{D_P}[A\mid\Xi^P=\xi].
\]

A single in-domain inverse risk only determines whether action information is
decodable within one transition domain. It cannot determine whether the learned
inverse relationship remains valid in the other domain. ATM therefore evaluates
each probe both in-domain and cross-domain:
\[
D_{i,j}
=
\mathbb{E}_{(\xi,a)\sim D_j}
\left[
\|h_i(\xi)-a\|_2^2
\right],
\qquad
i,j\in\{T,P\}.
\]

In particular, the pattern
\[
D_{P,P}\ll D_{P,T}
\]
shows that actions are highly decodable inside the model-predicted transition
domain while the corresponding inverse rule fails to transfer back to real
encoded transitions. This is precisely the predicted-domain-specific action
code discussed in the main paper.

Accordingly, ATM should be interpreted operationally as a cross-domain inverse
transfer test rather than as a direct estimator of a distributional divergence.
Its purpose is to determine whether an action--transition relationship learned
in one domain remains informative in the other.

The directional gaps
\[
G_{T\to P}
=
\frac{D_{T,P}-D_{T,T}}{D_{T,T}+\epsilon},
\qquad
G_{P\to T}
=
\frac{D_{P,T}-D_{P,P}}{D_{P,P}+\epsilon}
\]
summarize degradation in the two transfer directions. As in the main paper, we
treat $D_{T,T}$ as the primary diagnostic, cross-domain transfer as a secondary
diagnostic, and matrix-level asymmetry as a tertiary summary rather than a
standalone quality score.

\section{Additional Screening and Cross-Family Analysis}
\label{app:screening}

\subsection{Pairwise Ranking Protocol}
\label{app:ranking_protocol}

For OGBench-Cube, the screening score used in the main experiments is
\[
S_{\mathrm{ATM}}
=
-D_{T,T}
-
\lambda |G_{T\to P}|.
\]
The score is used only for relative model ranking rather than as a calibrated
predictor of absolute success.

For a pair of candidate models, ranking is counted as correct when the ordering
induced by $S_{\mathrm{ATM}}$ agrees with the ordering induced by
simulator-based planning success. Success-margin evaluation removes pairs whose
simulator success rates differ by less than the specified margin. This avoids
placing the same emphasis on nearly indistinguishable candidates and candidates
whose downstream performance differs substantially.

\subsection{Cross-Family Ranking}
\label{app:cross_family}

Table~\ref{tab:app_cross_family} provides the complete cross-family comparison
between LeWM-style~\citep{maes2026lewm} and DINO-WM~\citep{zhou2024dinowm}
candidates on OGBench-Cube. The transferred setting uses the coefficient fitted
only on LeWM-style candidates, whereas the refit setting calibrates the same
single coefficient on the mixed candidate pool.

\begin{table}[ht]
\centering
\caption{
Cross-family ranking on OGBench-Cube.
DINO-only ranking is included as a stress case because it contains only
40 pairs over a relatively narrow success range. The LeWM--DINO pairs are the
primary cross-family comparison.
}
\label{tab:app_cross_family}
\begin{tabular}{llccc}
\toprule
Setting & Pair type & $\lambda$ & $n_{\mathrm{pairs}}$ & Ranking acc. \\
\midrule
Transferred & DINO-only   & 0.036 & 40   & 70.00 \\
Transferred & LeWM--DINO  & 0.036 & 604  & 68.54 \\
Transferred & All pooled  & 0.036 & 2509 & 78.24 \\
\midrule
Refit       & DINO-only   & 0.005 & 40   & 32.50 \\
Refit       & LeWM--DINO  & 0.005 & 604  & 89.90 \\
Refit       & All pooled  & 0.005 & 2509 & 83.10 \\
\bottomrule
\end{tabular}
\end{table}

The transferred results show that direct score calibration does not completely
transfer across world-model families. However, refitting only the single
coefficient on the mixed candidate pool substantially improves LeWM--DINO
ranking. This supports the interpretation in the main paper that the underlying
ATM quantities remain informative across model families while their relative
scales can differ.

\subsection{Cross-Family Calibration Visualization}
\label{app:calibration}

\begin{figure}[ht]
    \centering
    \includegraphics[width=0.62\linewidth]{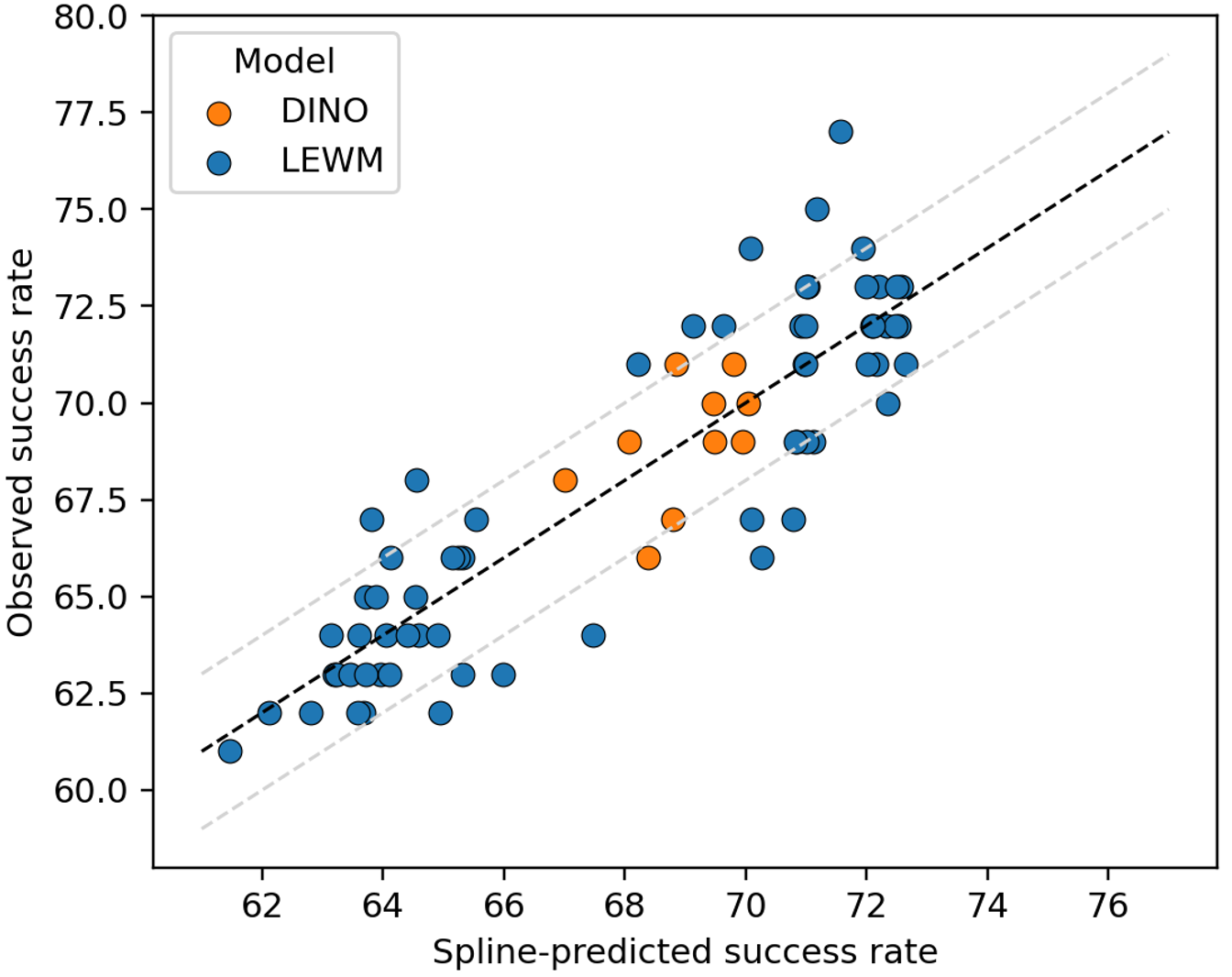}
    \caption{
    Cross-family calibration on OGBench-Cube. A low-capacity spline maps ATM
    diagnostics to downstream planning success for visualization. The
    calibration is not used as the primary screening score or for the main
    pairwise-ranking results.
    }
    \label{fig:app_calibration}
\end{figure}

Figure~\ref{fig:app_calibration} provides a complementary visualization of this
analysis. A low-capacity spline is fitted on the pooled candidate set only for
visualization; it is not used to define ATM or to produce the main pairwise
ranking results.

\section{Computational Cost}
\label{app:efficiency}

The screening experiments also differ substantially from the simulator-based
CEM evaluation used in this work in their wall-clock cost.
Table~\ref{tab:app_efficiency} reports the measured evaluation time under our
experimental setup. CEM evaluation requires repeated planner--model interaction
over simulator episodes, whereas ATM operates on cached latent transitions and
trains only the lightweight post-hoc probes.

\begin{table}[ht]
\centering
\caption{
Wall-clock cost under the evaluation protocol used in this work.
ATM timing is measured after latent transitions have been cached. These numbers
characterize the practical screening cost in our setup rather than an intrinsic
efficiency advantage over all possible planning methods.
}
\label{tab:app_efficiency}
\begin{tabular}{lccc}
\toprule
Protocol & Simulator rollout? & 50 episodes & 100 episodes \\
\midrule
LeWM CEM evaluation~\citep{maes2026lewm}
    & Yes & $\sim 3$ min & $\sim 7$ min \\
DINO-WM CEM evaluation~\citep{zhou2024dinowm}
    & Yes & $\sim 1.4$ h & $\sim 2.8$ h \\
ATM diagnostics
    & No & \multicolumn{2}{c}{3--5 s} \\
\bottomrule
\end{tabular}
\end{table}

Once latent transitions are cached, ATM requires only approximately
3--5 seconds per checkpoint in our setup. In comparison, the simulator-based
CEM evaluation used here requires minutes for LeWM-style models and hours for
DINO-WM. We therefore view computational efficiency as a practical benefit of
ATM-based screening under the present evaluation setting, rather than as the
defining contribution of ATM.
\end{document}